\pdfoutput=1

\documentclass[11pt]{article}
\usepackage{acl}

\usepackage{times}
\usepackage{latexsym}
\usepackage{xcolor}
\usepackage{graphicx}
\usepackage{caption}
\usepackage{subcaption}
\usepackage{soul}
\usepackage{booktabs}
\usepackage{amsmath}
\DeclareMathOperator*{\argmax}{arg\,max}

\usepackage[T1]{fontenc}
\usepackage{hyperref}
\usepackage {xcolor}
\usepackage{float}
\usepackage{makecell}

\usepackage[utf8]{inputenc}
\usepackage{tabularx}
    \newcolumntype{L}{>{\raggedright\arraybackslash}X}
\usepackage{microtype}
\title{A Dialogue System for Assessing Activities of Daily Living: Improving
Consistency with Grounded Knowledge}

\author{\textbf{Zhecheng Sheng \:\:\:\: Raymond Finzel \:\:\:\: Michael Lucke}\\
\textbf{Sheena Dufresne \:\:\:\: Maria Gini \:\:\:\: Serguei Pakhomov} \\
University of Minnesota. Twin Cities \\
\texttt{\{sheng136, finze006\}@umn.edu}
}

\begin{document}
\maketitle
\begin{abstract}

In healthcare, the ability to care for oneself is reflected in the "Activities of Daily Living (ADL)," which serve as a measure of functional ability (functioning). A lack of functioning may lead to poor living conditions requiring personal care and assistance. To accurately identify those in need of support, assistance programs continuously evaluate participants' functioning across various domains. However, the assessment process may encounter consistency issues when multiple assessors with varying levels of expertise are involved. Novice assessors, in particular, may lack the necessary preparation for real-world interactions with participants. To address this issue, we developed a dialogue system that simulates interactions between assessors and individuals of varying functioning in a natural and reproducible way. The dialogue system consists of two major modules, one for natural language understanding (NLU) and one for natural language generation (NLG), respectively. In order to generate responses consistent with the underlying knowledge base, the dialogue system requires both an understanding of the user's query and of biographical details of an individual being simulated. To fulfill this requirement, we experimented with query classification and generated responses based on those biographical details using some recently released InstructGPT-like models.
\end{abstract}

\section{Introduction}

Conversational AI is expanding beyond use in general applications like virtual assistants \citep{sciuto2018hey} to use in specialized domains such as healthcare and finance where it can aid patients or customers in various scenarios. Specifically, there is interest in applications of this technology for patient care and monitoring after hospital discharge \citep{fadhil2018beyond}. 
Assessing functioning is crucial in clinical and non-clinical fields, such as nursing, physical and occupational therapy, geriatric medicine, neurology, rheumatology, disability, and human services. A person's ability to perform day-to-day activities independently depends on their cognitive, motor, and perceptual abilities, which are collectively referred to as Activities of Daily Living (ADL) \citep{edemekong2022activities}. Impairments in these abilities often require assistive devices, external supervision, assistance, or a long-term support plan.
The Minnesota Department of Human Services (MNDHS) provides significant public resources to assist individuals with impaired functioning based on their specific needs. Certified assessors conduct face-to-face interviews with individuals to determine the level of support required to meet their needs, covering a wide range of areas related to ADLs. The goal of these assessments is to determine an individual's level of independence in performing ADLs. However, ensuring consistency across numerous assessors (e.g., ~1,700 in the state of Minnesota) and preparing novice assessors for diverse field interactions poses a challenge to the state's intake process.

Despite the availability of free corpora for training end-to-end neural models, most dialogue systems in healthcare are still rule-based \citep{laranjo2018conversational}. With the proliferation of neural models, there has been an increasing concern about the factual consistency of AI-powered applications. Factual consistency with a knowledge source, explicated in prior work as knowledge-grounding or attributability \citep{rashkin2022measuring}, refers to the ability of a model to generate responses that are accurate and consistent with the information present in a verified knowledge base. Knowledge grounding is particularly important in language models used for tasks that require accurate information, such as question-answering, dialogue systems, and chatbots \citep{honovich-etal-2022-true, tam2022evaluating, nan-etal-2021-improving}. 

To tackle these challenges and facilitate the training of certified assessors in conducting ADL assessments, we propose a coaching dialogue system presented in this paper. Specifically, our contributions can be summarized as follows: 1. We created a novel dataset for developing and evaluating dialogue systems focused on ADL assessments. 2. We compared several statistical models used for natural language understanding. 3. We experimented with several approaches to grounding language generation with large language models by using knowledge contained in a manually constructed knowledge base.


\section{Related Work}
Relevant previous work has been conducted on dialogue systems developed for use in healthcare settings \citep{jaffe-etal-2015-interpreting, llanos2015description, nirenburg2008language, laleye-etal-2020-french}. These dialogue systems simulate a virtual standardized patient to deliver healthcare education from structured encounter data and rely on matching algorithms to extract scripted answers. In contrast, the dialogue system we have created generates responses that are dependent on various synthetic profile characteristics and allows off-topic conversations, thus offering a greater degree of variability while still remaining grounded in the knowledge contained in the synthetic profile. This allows the system to simulate several possible patients with different attributes, with the goal of giving assessors a chance to practice their interview skills with simulated patients of different functioning levels and communication styles. Knowledge grounding using external knowledge graphs in combination with transformer models was previously explored \citep{lucke2023, liu2021kgbart, agarwal-etal-2021-knowledge, koncel-kedziorski-etal-2019-text}. Open domain conversational dataset and dialogue systems based on factual knowledge have also been developed \citep{dinan2018wizard, dziri2022faithdial}. In this paper we focus on the fine-tuning of InstructGPT-like models, and also on the combination of these fine-tuned models with a knowledge base of pre-written natural language facts using query classification and information retrieval via similarity matching.
We previously published on earlier versions of this system, referred to as Conversational Agent for Daily Living Assessment Coaching(CADLAC) \citep{Gaydhani2020Conversational}. These earlier versions as well as the current version of the system were deployed as a demonstration with a web-based interface \citep{finzel-etal-2021-conversational}. 

\begin{figure*}[htbp]
    \centering
    \includegraphics[width=1\textwidth]{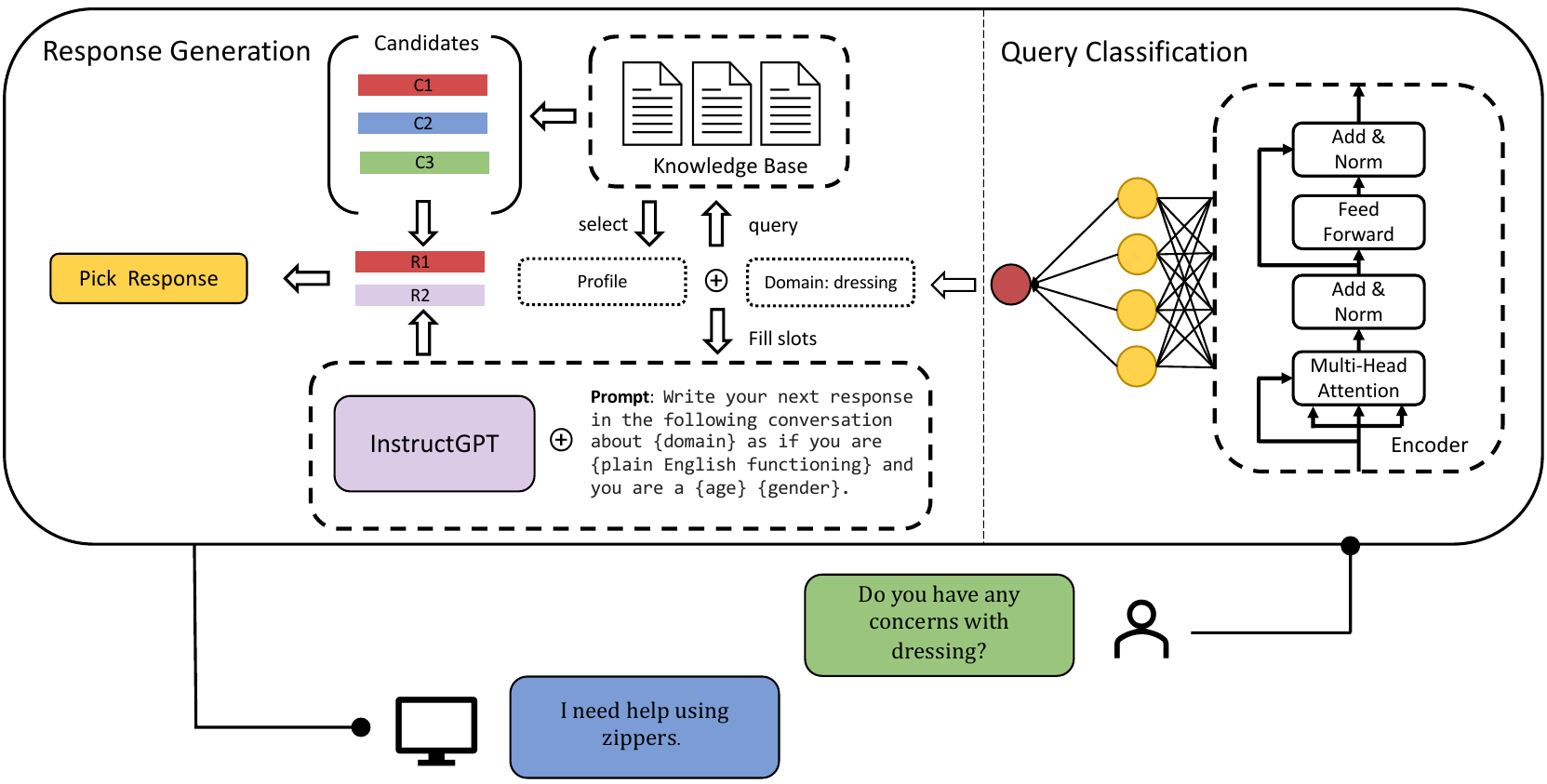}
    \caption{Workflow of the dialogue system: the user communicates with a pre-selected profile through the web interface and with typing or voice. A pre-trained classification model on the back-end dispatches the query to the correct domain. The system tries to match the query against the knowledge base through some similarity measurements. If there is no contents similar to the incoming query, it turns to a fine-tuned InstructGPT model to generate a reasonable turn of dialogue.}
    \label{fig1}
\end{figure*}

\section{Methods}
\subsection{Data}

\subsubsection{Synthetic Dialogues}
We administered a survey to approximately 1,700 certified assessors aiming to collect sample dialogues across 18 ADL domains(Appendix \ref{aa0}). The assessors were requested to recall interactions they had with participants during past assessment interviews and provide up to 3 dialogue turns between themselves and the person being interviewed. The survey also included questions on the gender and age category of the person, the domain of the conversation, and the person's ability level within the domain. The survey results in a total of 2,885 dialogues. A labeled sample record is shown in Appendix \ref{es}. The survey data was utilized to fine-tune a query classification model for our system.

\subsubsection{Historical Assessment}\label{histassess}
The grounded knowledge relies on the database of 10,000 historical assessments that were conducted by experienced certified assessors and were overseen by Minnesota Department of Human Services(DHS). Each assessment includes various fields that detail each individual's ability to perform ADLs, along with basic demographic information such as age range and gender. Additionally, the assessments contain notes taken by the certified assessors during the interview, which briefly describe the person's difficulties, preferences, and any assistive devices they use, among other information organized by the ADL domains. All historical records were anonymized by removing any individually identifiable information including names and exact age. Likewise, sensitive information such as phone numbers, email and physical addresses were removed. 

\subsubsection{Synthetic Profiles} 
The de-identified historical assessment notes were utilized to create synthetic profiles of individuals that specify varying levels of independence in 18 ADL domains and their specific needs. Categorical attributes related to independence levels in the historical assessments were mapped to numerical ratings to create these profiles. Furthermore, the synthetic profiles were populated with assessor notes about intents or action types from the historical data. Since the synthetic profiles were created based on real individual data, they can convey various sources of biases. To mitigate those biases, we conducted a stratified sampling based on gender and race from the original assessment collection, and create a balanced set of profiles in terms of demographics.

\subsubsection{Manual Annotation}\label{anno} 
The synthetic profiles are used as grounded documents for the dialogue system to generate responses that are tailored to the question asked by the assessor and are factual consistent with the underlying profile information. However, as the original historical assessments only represent brief descriptions of different ADL conditions for the assessed participant, they can not be directly used as materials for response generation. To overcome this challenge, we manually translated short assessor notes into natural conversations with several turns, which is correlated with the note. Specifically, our annotators, who had domain-related language expertise, wrote the responses by inferring what the person being assessed might have said during the assessment that led the assessor to jot down the particular note. To illustrate, suppose a 60-year old male was commented "Prefer shower" in the assessor's note. In that case, the annotator may deduce that during the interview, the assessed person answered \textit{"I do not like baths, I prefer to shower."} and continue the conversation with several follow-up responses: \textit{"I like taking long showers."}, \textit{"It's nice to have reminders to get out of the shower when I have been in there for a while."}

This proposed annotation guideline generated two distinct types of responses, "direct" and "indirect," based on the requirements of the assessment situation. Direct responses are written in first-person narrative, while indirect responses are in third-person narrative. The responses primarily rely on the information present in the assessor note, but other fields of information are also used to formulate the response. Direct speech is generated for adults, while indirect speech is used for simulating assessments of children, in which case the assessor would be interviewing the child's parents/caregivers. 

For the experiments presented in this paper, 10 annotated synthetic profiles were included and their characteristic are shown in Table \ref{profiles}. Given the amount of data we have collected and the efforts made to create the annotated conversations, we believe it would be a valuable novel corpus for the computational linguistics community. We plan to develop this corpus and release it in the near future.

\subsection{Dialogue System}
A typical dialogue system consists of distinct NLU and NLG modules that interact with a dialogue manager to maintain a conversation. For the rest of this paper we focus on the NLU and NLG modules which will be referred to as query classification and response generation, respectively, as shown in Fig \ref{fig1}. The back-end of the system is built upon the open conversational AI platform MindMeld \footnote{\url{https://github.com/cisco/mindmeld}}. The current iteration of the system is equipped with recently emerging deep transformer models, which represents a better ability to capture desired knowledge, and provides a framework in which to evaluate the capacity of the InstructGPT family of transformers.

\subsubsection{Query Classification}\label{nlu}
To ensure the factual consistency of the system, the incoming user query can be mapped to a domain and intent to assist the generative model in producing reasonable responses. In the following experimental setup, we only considered performance differences in the domain classification task because a large portion of intents under the same domain share similar utterances. When we apply similarity measurements we are searching over all the intents except certain ones(e.g. \textit{preference, equipment}). This strategy increases the system's sensitivity and makes the impact of intent classification more subtle. We used DistilBERT \citep{sanh2019distilbert} as the intent classifier for the sake of efficiency throughout the conversation experiments. 

\begin{figure}[ht]
  \centering
  \begin{subfigure}[b]{0.8\linewidth}
    \centering
    \includegraphics[width=1\linewidth]{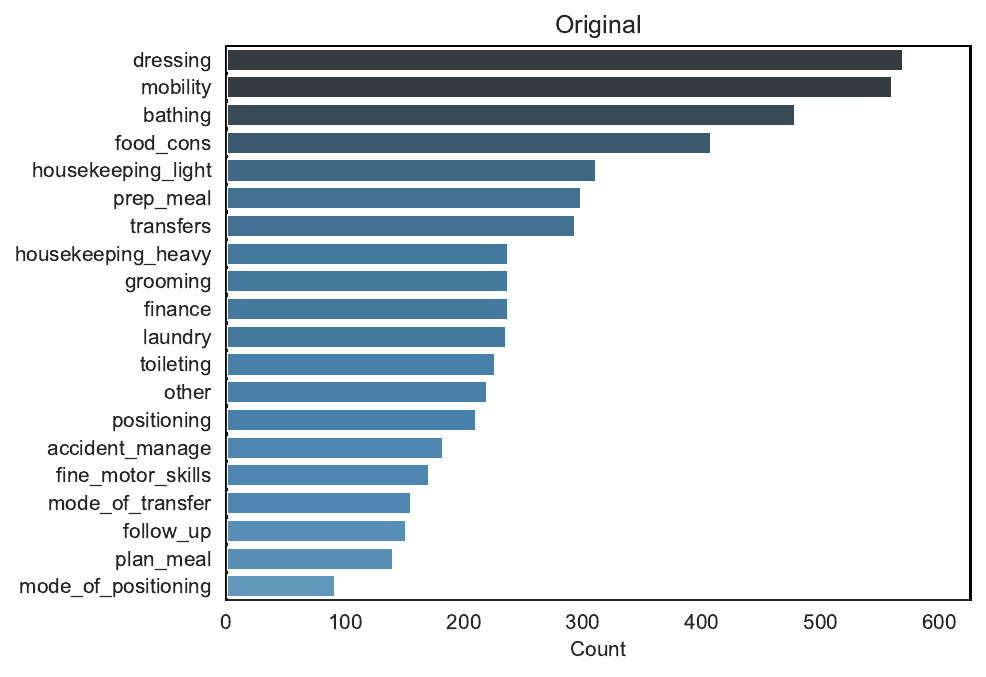}
    \label{ot}
  \end{subfigure}
  \hfill
  \begin{subfigure}[b]{0.8\linewidth}
    \centering
    \includegraphics[width=1\linewidth]{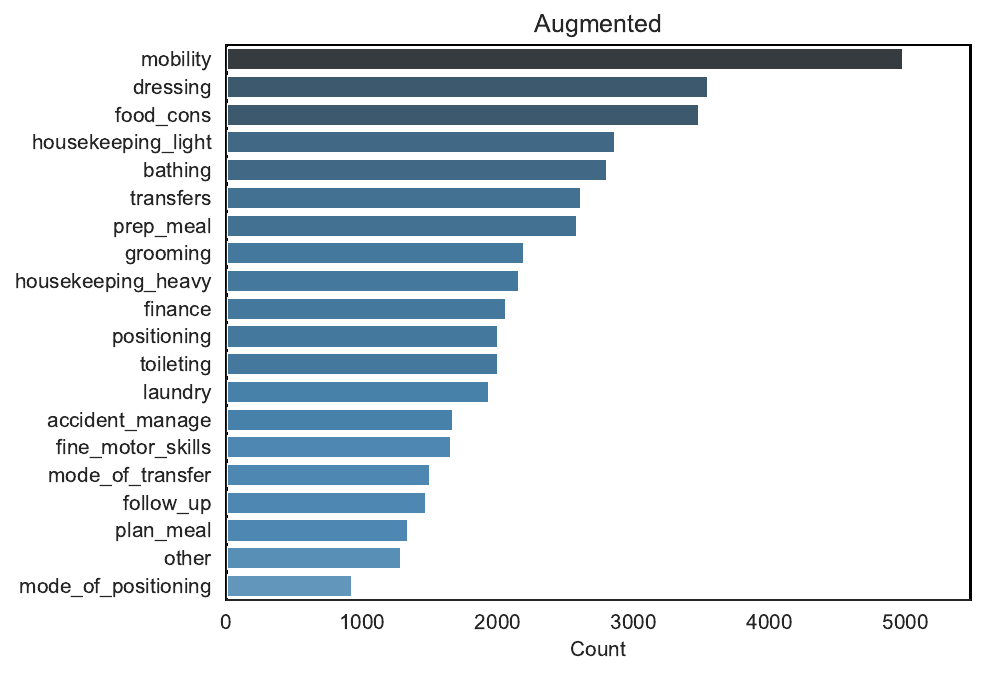}
    \label{at}
  \end{subfigure}
  \caption{Counts of examples for each domain in the training data.}
   \label{fig2}
\end{figure}

For domain classification we conducted experiments with 4 different models, ranging from simple multinomial logistic regression to transformer based encoders including BERT \citep{devlin2018bert}, RoBERTa \citep{conneau2019roberta} and DeBERTa \citep{he2021deberta}. All the pre-trained models from Section \ref{nlu} and \ref{nlg} were available from the public Huggingface model hub \footnote{\url{https://huggingface.co/models}}. In the deployment of the dialogue system, two additional fields \textit{follow up} and \textit{other} were added to existing domains to allow assessors to ask more about specific responses or to converse casually. A collection of phrases for greeting, ending the conversation, and also phrases for generic follow up questions were created and added to the labeled corpus specifically. For experiments, 20\% of the resultant corpus was randomly sampled for testing and the remaining 80\% was used for training. As one can observe from Figure \ref{fig2}, the original training corpus contains a limited number of utterances from each domain and may undermine the ability of the deep transformer models to distinguish between domains. To investigate the effect of this limited corpus, we applied the abstractive text summarization model PEGASUS \citep{zhang2019pegasus} to generate paraphrases of the original utterances. This model employed a self-supervised objective for pre-training transformer encoder-decoder models that involves removing several whole sentences from a document and then asking the model to recover them without extensive human annotation efforts. In our case, the model recovers and recreates utterances in the training corpus under each domain. The frequency distribution of the augmented training corpus is also shown in Figure \ref{fig2}.

We fine-tuned different models with the two training corpus and repeated each experiment 10 times to get the statistics of different settings.

\subsubsection{Response Generation}\label{nlg}
The primary goal of developing this dialogue system is to generate human-like responses that are consistent with factual information present in the knowledge base. This requires the generation to at least partially rely on the documents that have been collected and used to construct synthetic profiles. Even though there is plenty of evidence showing that large language models learn some factual knowledge during pre-training \citep{wang2020language} and could potentially be used as sources of accurate information \citep{petroni-etal-2019-language}, model adaptation is still needed for the models to represent concepts from specific domains. Model fine-tuning is a common way to ensure that a language model includes some external knowledge. This section will showcase assessments of the response generation of some InstructGPT-like models, including an evaluation of a zero-shot methodology, a fine-tuned model, and an assessment of a methodology that uses an InstructGPT-like model as a fallback when bespoke responses that are significantly similar to the current query (as determined by query classification) are not available in the knowledge base.

Over the past few months, large language models (LLMs) such as ChatGPT \citep{ouyang2022training}, have been garnering attention due to their impressive ability to understand instructions and generate human-like responses. The InstructGPT-like models are trained on massive amounts of natural language data in auto-regressive fashion and then fine-tuned to follow large-scale human instructions \citep{wang2022selfinstruct}. They exhibit robust performance across a wide range of natural language processing tasks and can generalize to unseen tasks, making them promising unified solutions for text generation and conversational AI. 

We hope to leverage the strength of open-source LLMs to generate answers by understanding assessor questions and responding with human-like reasoning about functioning. However, given that publicly available models are generally pre-trained on data outside of the ADL domains, it is necessary to create a dataset explicitly for our task. By further fine-tuning an InstructGPT on our ADL specific dataset, we can benefit from the model's conversation capability while also adapting to the style of an assessment interview. Researchers recently found that achieving the best performance with a fixed computer budget does not solely depend on model size. In some cases, smaller models that have been pre-trained with a greater amount of data can outperform larger models \citep{hoffmann2022training}. This is important to the deployment of applications like dialogue systems in the real world as they need to interact with users with very low latency. This requires models to have high computation efficiency at inference time. LLaMA \citep{touvron2023llama} is a set of fundamental instruct-based language models, varying in size from 7 billion to 65 billion parameters. The models were trained on a mixture data source consists of roughly 1.4 trillion unique tokens.
It has been reported LLaMA 7B models demonstrated competitive performances against GPT-3 on multiple tasks such as CommonSense Reasoning and Closed-book Question Answering. \citep{touvron2023llama}. Considering the computational burden, we decided to investigate the 7B LLaMA model and experiment under several settings to evaluate its factual consistency with the grounded documents.

As described earlier in Section \ref{anno}, the historical assessments were transformed into numbers of synthetic profiles with certain age, gender and various levels of daily living functioning in different domains. We rely on these synthetic profiles to establish the knowledge base, which currently contains 10 sampled profiles.

 When InstructGPT models are used in the dialogue system, it is essential to feed the model with a well-designed prompt which embeds factual context and also provides a clear description of the task. We first translated numerical ratings of functioning for each assessment into plain English, then inserted that information into tailored prompt templates. We designed one template for typical interrogative sentences and another for follow up questions.
\begin{enumerate}
    \item General: Write your next response in the following conversation about \texttt{\{domain\}} as if you \texttt{\{plain English functioning\}} and you are \texttt{\{age\} \{gender\}}.
    \item Follow-up: Provide more details to this statement about  \texttt{\{domain\}} as if you \texttt{\{plain English functioning\}} and you are \texttt{\{age\} \{gender\}}.
\end{enumerate}

\begin{figure}[htbp]
  \centering
    \includegraphics[width=0.7\linewidth]{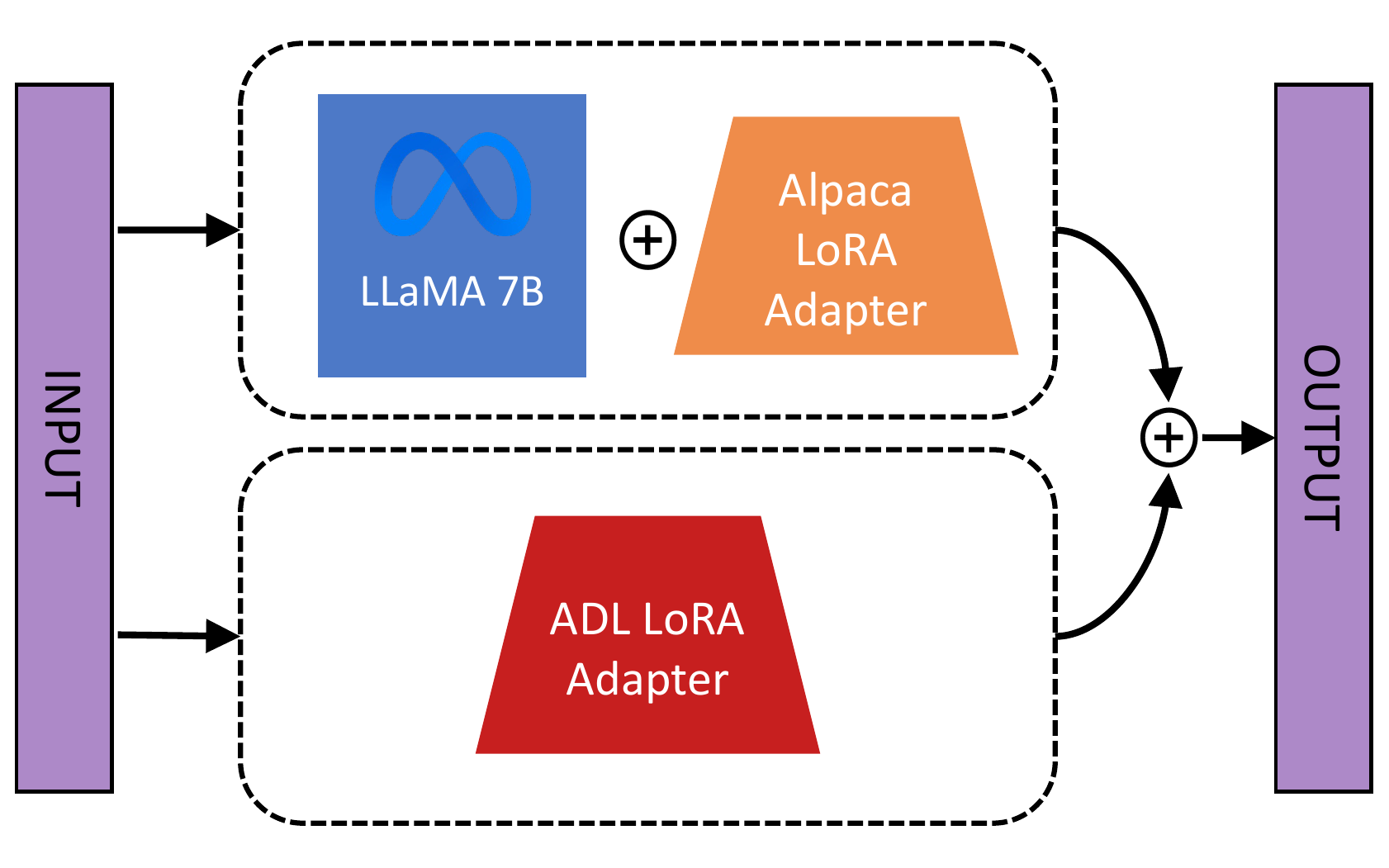}
    \caption{Diagram of LoRA training}
    \label{lora}
\end{figure}

 The fine-tuning data was derived from both the human written synthetic dialogues from survey data and annotated historical assessments introduced above. The instruction-following fine-tuning data format contains 3 fields: context, input, and output. The context field is filled with one of the prompts above, while the input and output fields are filled with one question answer pair. To accommodate multi-turn conversation from the source data, we concatenated all previous turns with a newline separator "\texttt{\textbackslash n}" to account for dialogue history. The resultant dataset has 6,123 question/answer pairs and examples of short conversations. The dataset covers diverse profiles and questions from all the ADL domains of interest. While fine-tuning the whole 7B models is prohibitively costly, there exist a family of methods called Parameter Efficient Fine-Tuning methods(PEFT) that only train a tiny number of parameters, but which result in comparable performance to whole-model fine-tuning \citep{peft}. Low-Rank Adaptation(LoRA) \citep{hu2021lora} is the one of those methods we selected for our task. The idea behind LoRA is to fix the pre-trained model weights and add trainable rank decomposition matrices to every layer of the transformer architecture. In order to train the LoRA adapter with conversational capabilities for activities of daily living context, we apply the strategy demonstrated in Figure \ref{lora}. We first trained a LoRA adapter using the public Stanford Alpaca dataset \citep{alpaca} to take advantage of the 52K instruction-following dataset, and then merged the LoRA weights back into LLaMA to create a single base model. We then trained another LoRA adapter with our ADL conversation specific instruction-following dataset described above. This training strategy is similar to the one employed in ChatDoctor \citep{yunxiang2023chatdoctor}, but at this time we have only investigated the LoRA training approach, while ChatDoctor did a concurrent investigation of fine-tuning the whole model with the Stanford Alpaca dataset. All of our locally trained models were quantized using 8-bit precision to allow for fine-tuning on a single GPU.

\begin{table*}[htbp]
\centering
\resizebox{\linewidth}{!}{%
\begin{tabular}{l|l|l|l|l}
\toprule
\textbf{Experiments} & \textbf{Accuracy} & \textbf{F1-weighted} & \textbf{F1-micro} &\textbf{F1-macro}\\
\midrule
LR + Original & $0.703_{(0.702-0.704)}$ & $0.708_{(0.707-0.709)}$ & $0.703_{(0.702-0.704)}$ & $0.606_{(0.604-0.608)}$ \\
LR + Augmented & $0.696_{(0.694-0.705)}$ & $0.702_{(0.700-0.711)}$ & $0.696_{(0.694-0.705)}$ & $0.615_{(0.613-0.623)}$ \\
BERT$_{base}$ + Original & $0.747_{(0.729-0.760)}$ & $0.744_{(0.727-0.756)}$ & $0.747_{(0.729-0.760)}$ & $0.649_{(0.635-0.670)}$ \\
BERT$_{base}$ + Augmented & $0.726_{(0.720-0.733)}$ & $0.729_{(0.723-0.738)}$ & $0.726_{(0.720-0.733)}$ & $0.639_{(0.630-0.651)}$ \\
RoBERTa$_{base}$ + Original & $0.759_{(0.745-0.767)}$ & $0.757_{(0.740-0.766)}$ & $0.759_{(0.745-0.767)}$ & $0.667_{(0.629-0.698)}$ \\
RoBERTa$_{base}$ + Augmented & $0.727_{(0.720-0.732)}$ & $0.731_{(0.725-0.737)}$ & $0.727_{(0.720-0.732)}$ & $0.641_{(0.633-0.648)}$ \\
DeBERTa$_{v3}$ + Original & $\mathbf{0.762}_{(0.752-0.782)}$ & $\mathbf{0.759}_{(0.746-0.781)}$ & $\mathbf{0.762}_{(0.752-0.782)}$ &$\mathbf{0.683}_{(0.652-0.708)}$ \\
DeBERTa$_{v3}$ + Augmented & $0.732_{(0.728-0.738)}$ & $0.736_{(0.732-0.741)}$ & $0.732_{(0.728-0.738)}$ & $0.646_{(0.643-0.651)}$ \\
\bottomrule
\end{tabular}%
}
\caption{\label{expdomain}
Experimental results of testing classification models. The best performer for each metric is marked in bold.
}
\end{table*}

\subsection{Evaluation Methods}
\subsubsection{Query Classification} 

In order to generate sensible responses from our knowledge base of pre-prepared facts, the domain of a given query needed to be classified so that the system could accurately identify candidate responses. We assessed domain classification performance, across 4 different metrics, including weighted $F_1$, micro $F_1$ and macro $F_1$ for multi-classification task. The accuracy measures how the model performs regardless of the domain differences and the other 3 aggregated f-measures implies the performance when imbalance exists across domains.
\subsubsection{Response Generation}
In dialogue system research, evaluating the quality of the conversation automatically is still an open problem \citep{Deriu2020}. There have been efforts to develop reference-free metrics for evaluating factual consistency in knowledge grounded dialogue systems \citep{honovich-etal-2021-q2} based on automatic question generation paired with a question answering model. However, given the style of the knowledge base in our system and various possible definitions of factuality, we only pursued human ratings at this time to evaluate the quality of conversations. We evaluated text excerpts through the notion of sensibleness and specificity, and provided a separate evaluation of factual consistency.
Sensibleness and specificity average (SSA) is a metric to capture human likeness of generated responses  \citep{adiwardana2020humanlike}. Sensibleness measures whether the generated response is coherent and makes sense given the context while specificity measures whether the generated response seems uniquely suited to the questions that are asked, rather than just sensible in general. We generated a short conversation snippet using each NLG method with a fixed set of questions in an effort to keep the style of conversation consistent across methods. Two domains for which we had the most data (bathing, dressing) were selected and we created 5 questions for each. The 5 questions comprise 1 general question, 1 follow up question and 3 questions for detailed aspects. Next we randomly selected one profile and filled those questions into the prompt by design. Besides our fine-tuned LLaMA model, we also tested a 13B Vicuna model \citep{vicuna2023} with a zero-shot configuration. In total, three models (Fine-tuned 7B LLaMA only, Fine-tuned 7B LLaMA + knowledge base, 13B Vicuna demo\footnote{\url{https://chat.lmsys.org}}) were accessed to generate conversations. When combining the fine-tuned LLaMA model with grounded knowledge, we used a heuristic rule to determine whether to utilize knowledge directly from the knowledge base, or to generate a response using the LLM. Mathematically, the heuristic can be expressed as:
\[
    R =
    \begin{cases}
      \underset{c \in \mathcal{C}}{\argmax} \: \sigma(q, c)  & \text{if } \underset{c \in \mathcal{C}}{\max} \: \sigma(q, c) \geq \lambda \\
      r_l & \text{otherwise}
    \end{cases}
\]
Where $\sigma$ is any similarity measurement(e.g. Bertscore), $\lambda$ is an arbitrary cutoff and $\mathcal{C}$ is the collection of all candidates from the knowledge base. $q$ denotes the incoming query and $r_l$ denotes the response generated from LLM.  A note about generation: in most cases, the fine-tuned LLaMA model tended to generate complete conversations rather than single responses, which could be due to the fact that the Alpaca fine-tuning data does not represent an obvious conversational form. To mitigate this, we manually selected the first sentence from the entire output as the response to the question. After assembling these excerpts we asked 6 colleagues who have limited background of this project to score sensibleness and specificity for each conversation on a scale of 1-6 and had them pick their favorite conversations based both on realistic quality and personal preference. The \href{https://docs.google.com/forms/d/e/1FAIpQLSe3QYV65nEGEDaoH-nU1lHEpwFInKAiz0UAjcYib_vnIyC5Rg/viewform}{evaluation form} is also publicly available. This resulted in 12 total SSA ratings of each method (2 conversation snippets per method, 6 raters) and 6 opinions on total quality and reality. For our factual consistency evaluation, two co-authors conversed freely with the chatbot systems for a fixed number of dialogue turns. This provided a chat history in which the dialogue systems would be free to "hallucinate" factual content about the synthetic profile, or forget about details that were already present in the conversation several turns ago. This method was selected for conversation generation in order to assure that the human turns were natural---allowing for things like specific follow-up questions, requests for further information, attempts to repair disfluent conversation turns, and other intricacies of human conversation. The co-authors then counted the number of contradictions that the conversations made against the knowledge base and also counted the number of factual self-contradictions in the dialogue history. This combination of two sorts of human-rated metrics (sensibility \& specificity, and external grounding \& internal consistency) formed our baseline for evaluating a systems ability to respond fluently and factually to an assessor's queries.

\begin{table*}[htbp]
\centering
\resizebox{\linewidth}{!}{%
\begin{tabular}{lcccccc}
\toprule
\textbf{Model} & Sensibleness & Specificity & Realness&Favorite & $\neg$ Knowledge & $\neg$ History \\
\midrule
Fine-tuned LLaMA 7B & 3.67 & 3.92 & 1 & 1 & 4 & 1\\
Zero shot Vicuna 13B & 4.50 & 5.00 & 0 & 1 & 5 & 2\\
Fine-tuned LLaMA 7B + Knowledge base & 4.92 & 4.33 & 5 & 4 & 1 & 0\\
\bottomrule
\end{tabular}%
}
\caption{\label{systemeval}
Human Evaluation Results. The numbers in the \textit{Sensibleness} and \textit{Specificity} columns represent the average rating across evaluators. Numbers in the remaining columns are simple counts. "$\neg$" indicates a contradiction against an existing knowledge source.
}
\end{table*}
\section{Results}
\subsection{Query Classification}
The experimental results for 4 different metrics are reported in Table \ref{expdomain}. We show the range of each metric instead of standard deviation as the numbers are too small compared to the mean value. The results indicate that the transformer family outperforms the simple logistic regression model with bag of word features.  DeBERTa$_{v3}$, the largest transformer model among the candidates, achieves the best performance for all 4 metrics. When comparing model training results between the training with an augmented corpus and training with the original corpus, we observe that all the models consistently perform better with the original corpus. This finding can indicate that larger quantities of data does not necessarily bring advantages to learning classification rules and we suspect the paraphrasing potentially introduces noise that lowers the quality of the training corpus. Comparisons across the 4 metrics suggests there is an imbalance in performance across the domains. Higher weighted $F_1$ score and micro $F_1$ score than macro $F_1$ score implies the model performs poorly on domains with less available data. Such performance imbalances can also arise when domains share a high degree of overlap in their conceptual definition (such as categories like \textit{light housekeeping} and \textit{heavy housekeeping}), or when there are differences in the size or variability of the available data. If interested, a per-domain breakdown of $F_1$ score can be found in Figure \ref{domainf1} (appendix).


\subsection{Response Generation}

Human functioning is not easily reduceable to an array of numbers, so grounding the knowledge in a way that respects the "functional levels" of the ADL, but also embeds knowledge of specific human details that differ from person to person is a challenge. In our evaluation of response generation using InstructGPT-like models, the knowledge-grounding process that we employed had a modest impact on system's ability to speak fluently and to speak into topics in which we did not have our own training data, such as regular open-domain conversation and non-functioning related conversation topics about home life. In the authors' opinion the open-domain response generation of raw LLMs provided a more pleasurable chat experience across a long conversation (something that our numeric evaluation across five and ten turn excerpts could not capture), but the tradeoff for factual and internal consistency provides value in the application of these technologies for the simulation of a factually grounded profile. The results in Table \ref{systemeval} indicate that despite the immense power of LLMs, facts stored as natural language snippets in a database may be used to improve factual and internal consistency, and this does not come at a penalty to a simulation's sensibleness, specificity, or realistic behavior in all cases. It is also interesting to note that when evaluating our LLM generated responses, we experimented with different hyperparameters and found that though the models provided different occasionally during this exploration, the knowledge conveyed from each run was consistent. For example, it was unlikely in our experience that a change to the decoding hyperparameters would cause the LLM to generate "I have no problem bathing on my own," when under another hyperparameter configuration it had responded "I need a lot of help with bathing."



\section{Conclusions and Limitations}
In this paper, we present a comprehensive framework for measuring the quality of a dialogue system dedicated to activities of daily living assessments. We have created a new high-quality dataset of human-written questions and answers with corresponding profile information. We are currently working on expanding the dataset by adding more profiles and removing any factual inconsistencies resulting from human error.
Although more complex models showed better query classification performances, we need to consider the trade-offs between model size and generation time in the deployment environment to ensure a smooth user experience. We also identify areas where LLM performance can be augmented by a knowledge base filled with human written natural language facts, and that this augmentation need not come at a penalty to sensibleness, specificity, or the realistic quality of conversation.
General conclusions based on our initial work here may not be possible given the limited number of evaluators and small amount of evaluated dialogues, and this is a major limitation of our contribution. Future work is needed to develop a more robust and replicatable evaluation framework, especially to perform evaluations of long and complex conversations like the type that assessors perform in the field. Such an evaluation will need to include larger numbers of human raters to improve the statistical power of the surveys. Recent automatic evaluations may also help improve development efforts, as a sufficiently powerful LLM such as GPT-4 may be able to monitor the chatbot for regressions in its ability to speak fluently, sensibly or specifically. This assessment, known informally as the "Vicuna Assessment"\citep{vicuna2023}, cannot give an evaluation of the chatbot's fit-for-purpose, but could be used to compare short conversations from several versions of the same chatbot. This could free up more human resources to evaluate the knowledge-groundedness and fit-for-purpose of future versions.
In addition, given more computing budget and more time to engineer prompts, larger language models beyond LLaMA 7B could be further studied or fine-tuned while experimenting with fine-tuning datasets and process. There are also many thresholds and parameters that could be further tested in the development of the knowledge-grounding system, wherein similarity measures inform the system's decision to answer using a generative model versus responding with language directly from the knowledge base.

\section{Acknowledgements}
Work on this paper was supported by funding from the Minnesota Department of Human Services. We would like to thank the people at DSD and MNIT for help with project specifications, gathering of historical data, and expert guidance on domain-specific aspects of the project. We would also like to thank Pamela Miller, Sidney Kiltie, and Elise Moore for help with transforming certified assessor notes to natural language format and Julia Garbuz for helping to develop and conduct the surveys of DHS assessors.

\bibliography{anthology,custom}

\begin{thebibliography}{39}
\expandafter\ifx\csname natexlab\endcsname\relax\def\natexlab#1{#1}\fi

\bibitem[{Adiwardana et~al.(2020)Adiwardana, Luong, So, Hall, Fiedel,
  Thoppilan, Yang, Kulshreshtha, Nemade, Lu, and Le}]{adiwardana2020humanlike}
Daniel Adiwardana, Minh-Thang Luong, David~R. So, Jamie Hall, Noah Fiedel,
  Romal Thoppilan, Zi~Yang, Apoorv Kulshreshtha, Gaurav Nemade, Yifeng Lu, and
  Quoc~V. Le. 2020.
\newblock \href {http://arxiv.org/abs/2001.09977} {Towards a human-like
  open-domain chatbot}.

\bibitem[{Agarwal et~al.(2021)Agarwal, Ge, Shakeri, and
  Al-Rfou}]{agarwal-etal-2021-knowledge}
Oshin Agarwal, Heming Ge, Siamak Shakeri, and Rami Al-Rfou. 2021.
\newblock \href {https://doi.org/10.18653/v1/2021.naacl-main.278} {Knowledge
  graph based synthetic corpus generation for knowledge-enhanced language model
  pre-training}.
\newblock In \emph{Proceedings of the 2021 Conference of the North American
  Chapter of the Association for Computational Linguistics: Human Language
  Technologies}, pages 3554--3565, Online. Association for Computational
  Linguistics.

\bibitem[{Chiang et~al.(2023)Chiang, Li, Lin, Sheng, Wu, Zhang, Zheng, Zhuang,
  Zhuang, Gonzalez, Stoica, and Xing}]{vicuna2023}
Wei-Lin Chiang, Zhuohan Li, Zi~Lin, Ying Sheng, Zhanghao Wu, Hao Zhang, Lianmin
  Zheng, Siyuan Zhuang, Yonghao Zhuang, Joseph~E. Gonzalez, Ion Stoica, and
  Eric~P. Xing. 2023.
\newblock \href {https://vicuna.lmsys.org} {Vicuna: An open-source chatbot
  impressing gpt-4 with 90\%* chatgpt quality}.

\bibitem[{Conneau et~al.(2019)Conneau, Khandelwal, Goyal, Chaudhary, Wenzek,
  Guzm{\'{a}}n, Grave, Ott, Zettlemoyer, and Stoyanov}]{conneau2019roberta}
Alexis Conneau, Kartikay Khandelwal, Naman Goyal, Vishrav Chaudhary, Guillaume
  Wenzek, Francisco Guzm{\'{a}}n, Edouard Grave, Myle Ott, Luke Zettlemoyer,
  and Veselin Stoyanov. 2019.
\newblock \href {http://arxiv.org/abs/1911.02116} {Unsupervised cross-lingual
  representation learning at scale}.
\newblock \emph{CoRR}, abs/1911.02116.

\bibitem[{Deriu et~al.(2020)Deriu, Rodrigo, Otegi, Echegoyen, Rosset, Agirre,
  and Cieliebak}]{Deriu2020}
Jan Deriu, Alvaro Rodrigo, Arantxa Otegi, Guillermo Echegoyen, Sophie Rosset,
  Eneko Agirre, and Mark Cieliebak. 2020.
\newblock \href {https://doi.org/10.1007/s10462-020-09866-x} {Survey on
  evaluation methods for dialogue systems}.
\newblock \emph{Artificial Intelligence Review}, 54(1):755--810.

\bibitem[{Devlin et~al.(2018)Devlin, Chang, Lee, and
  Toutanova}]{devlin2018bert}
Jacob Devlin, Ming{-}Wei Chang, Kenton Lee, and Kristina Toutanova. 2018.
\newblock \href {http://arxiv.org/abs/1810.04805} {{BERT:} pre-training of deep
  bidirectional transformers for language understanding}.
\newblock \emph{CoRR}, abs/1810.04805.

\bibitem[{Dinan et~al.(2019)Dinan, Roller, Shuster, Fan, Auli, and
  Weston}]{dinan2018wizard}
Emily Dinan, Stephen Roller, Kurt Shuster, Angela Fan, Michael Auli, and Jason
  Weston. 2019.
\newblock \href {https://openreview.net/forum?id=r1l73iRqKm} {Wizard of
  wikipedia: Knowledge-powered conversational agents}.
\newblock In \emph{International Conference on Learning Representations}.

\bibitem[{Dziri et~al.(2022)Dziri, Kamalloo, Milton, Zaiane, Yu, Ponti, and
  Reddy}]{dziri2022faithdial}
Nouha Dziri, Ehsan Kamalloo, Sivan Milton, Osmar Zaiane, Mo~Yu, Edoardo~M
  Ponti, and Siva Reddy. 2022.
\newblock \href {https://doi.org/10.1162/tacl_a_00529} {{FaithDial: A Faithful
  Benchmark for Information-Seeking Dialogue}}.
\newblock \emph{Transactions of the Association for Computational Linguistics},
  10:1473--1490.

\bibitem[{Edemekong et~al.(2023)Edemekong, Bomgaars, Sukumaran, and
  Schoo}]{edemekong2022activities}
Precious~F Edemekong, Dana~L Bomgaars, Shaji Sukumaran, and Christopher Schoo.
  2023.
\newblock \href {https://www.ncbi.nlm.nih.gov/books/NBK470380/}
  {\emph{Activities of Daily Living}}.
\newblock StatPearls Publishing.
\newblock PMID: 29261878.

\bibitem[{Fadhil(2018)}]{fadhil2018beyond}
Ahmed Fadhil. 2018.
\newblock \href {http://arxiv.org/abs/1803.06000} {Beyond patient monitoring:
  Conversational agents role in telemedicine and healthcare support for
  home-living elderly individuals}.
\newblock \emph{arXiv preprint arXiv:1803.06000}.

\bibitem[{Finzel et~al.(2021)Finzel, Gaydhani, Dufresne, Gini, and
  Pakhomov}]{finzel-etal-2021-conversational}
Raymond Finzel, Aditya Gaydhani, Sheena Dufresne, Maria Gini, and Serguei
  Pakhomov. 2021.
\newblock \href {https://doi.org/10.18653/v1/2021.eacl-demos.38}
  {Conversational agent for daily living assessment coaching demo}.
\newblock In \emph{Proceedings of the 16th Conference of the European Chapter
  of the Association for Computational Linguistics: System Demonstrations},
  pages 321--328, Online. Association for Computational Linguistics.

\bibitem[{Gaydhani et~al.(2020)Gaydhani, Finzel, Dufresne, Gini, and
  Pakhomov}]{Gaydhani2020Conversational}
Aditya Gaydhani, Raymond Finzel, Sheena~R Dufresne, Maria Gini, and Serguei
  Pakhomov. 2020.
\newblock Conversational agent for daily living assessment coaching.
\newblock \emph{CEUR Workshop Proceedings}, 2760:8--13.

\bibitem[{He et~al.(2021)He, Liu, Gao, and Chen}]{he2021deberta}
Pengcheng He, Xiaodong Liu, Jianfeng Gao, and Weizhu Chen. 2021.
\newblock \href {https://openreview.net/forum?id=XPZIaotutsD} {Deberta:
  Decoding-enhanced bert with disentangled attention}.
\newblock In \emph{International Conference on Learning Representations}.

\bibitem[{Hoffmann et~al.(2022)Hoffmann, Borgeaud, Mensch, Buchatskaya, Cai,
  Rutherford, de~Las~Casas, Hendricks, Welbl, Clark, Hennigan, Noland,
  Millican, van~den Driessche, Damoc, Guy, Osindero, Simonyan, Elsen, Rae,
  Vinyals, and Sifre}]{hoffmann2022training}
Jordan Hoffmann, Sebastian Borgeaud, Arthur Mensch, Elena Buchatskaya, Trevor
  Cai, Eliza Rutherford, Diego de~Las~Casas, Lisa~Anne Hendricks, Johannes
  Welbl, Aidan Clark, Tom Hennigan, Eric Noland, Katie Millican, George van~den
  Driessche, Bogdan Damoc, Aurelia Guy, Simon Osindero, Karen Simonyan, Erich
  Elsen, Jack~W. Rae, Oriol Vinyals, and Laurent Sifre. 2022.
\newblock Training compute-optimal large language models.
\newblock \emph{arXiv preprint arXiv:2202.09743}.

\bibitem[{Honovich et~al.(2022)Honovich, Aharoni, Herzig, Taitelbaum,
  Kukliansy, Cohen, Scialom, Szpektor, Hassidim, and
  Matias}]{honovich-etal-2022-true}
Or~Honovich, Roee Aharoni, Jonathan Herzig, Hagai Taitelbaum, Doron Kukliansy,
  Vered Cohen, Thomas Scialom, Idan Szpektor, Avinatan Hassidim, and Yossi
  Matias. 2022.
\newblock \href {https://doi.org/10.18653/v1/2022.dialdoc-1.19} {{TRUE}:
  Re-evaluating factual consistency evaluation}.
\newblock In \emph{Proceedings of the Second DialDoc Workshop on
  Document-grounded Dialogue and Conversational Question Answering}, pages
  161--175, Dublin, Ireland. Association for Computational Linguistics.

\bibitem[{Honovich et~al.(2021)Honovich, Choshen, Aharoni, Neeman, Szpektor,
  and Abend}]{honovich-etal-2021-q2}
Or~Honovich, Leshem Choshen, Roee Aharoni, Ella Neeman, Idan Szpektor, and Omri
  Abend. 2021.
\newblock \href {https://doi.org/10.18653/v1/2021.emnlp-main.619} {$q^{2}$:
  {E}valuating factual consistency in knowledge-grounded dialogues via question
  generation and question answering}.
\newblock In \emph{Proceedings of the 2021 Conference on Empirical Methods in
  Natural Language Processing}, pages 7856--7870, Online and Punta Cana,
  Dominican Republic. Association for Computational Linguistics.

\bibitem[{Hu et~al.(2021)Hu, Shen, Wallis, Allen-Zhu, Li, Wang, Wang, and
  Chen}]{hu2021lora}
Edward~J. Hu, Yelong Shen, Phillip Wallis, Zeyuan Allen-Zhu, Yuanzhi Li, Shean
  Wang, Lu~Wang, and Weizhu Chen. 2021.
\newblock \href {http://arxiv.org/abs/2106.09685} {Lora: Low-rank adaptation of
  large language models}.

\bibitem[{Jaffe et~al.(2015)Jaffe, White, Schuler, Fosler-Lussier, Rosenfeld,
  and Danforth}]{jaffe-etal-2015-interpreting}
Evan Jaffe, Michael White, William Schuler, Eric Fosler-Lussier, Alex
  Rosenfeld, and Douglas Danforth. 2015.
\newblock \href {https://doi.org/10.3115/v1/W15-0611} {Interpreting questions
  with a log-linear ranking model in a virtual patient dialogue system}.
\newblock In \emph{Proceedings of the Tenth Workshop on Innovative Use of {NLP}
  for Building Educational Applications}, pages 86--96, Denver, Colorado.
  Association for Computational Linguistics.

\bibitem[{Koncel-Kedziorski et~al.(2019)Koncel-Kedziorski, Bekal, Luan, Lapata,
  and Hajishirzi}]{koncel-kedziorski-etal-2019-text}
Rik Koncel-Kedziorski, Dhanush Bekal, Yi~Luan, Mirella Lapata, and Hannaneh
  Hajishirzi. 2019.
\newblock \href {https://doi.org/10.18653/v1/N19-1238} {{T}ext {G}eneration
  from {K}nowledge {G}raphs with {G}raph {T}ransformers}.
\newblock In \emph{Proceedings of the 2019 Conference of the North {A}merican
  Chapter of the Association for Computational Linguistics: Human Language
  Technologies, Volume 1 (Long and Short Papers)}, pages 2284--2293,
  Minneapolis, Minnesota. Association for Computational Linguistics.

\bibitem[{Laleye et~al.(2020)Laleye, de~Chalendar, Blani{\'e}, Brouquet, and
  Behnamou}]{laleye-etal-2020-french}
Fr{\'e}jus A.~A. Laleye, Ga{\"e}l de~Chalendar, Antonia Blani{\'e}, Antoine
  Brouquet, and Dan Behnamou. 2020.
\newblock \href {https://aclanthology.org/2020.lrec-1.72} {A {F}rench medical
  conversations corpus annotated for a virtual patient dialogue system}.
\newblock In \emph{Proceedings of the Twelfth Language Resources and Evaluation
  Conference}, pages 574--580, Marseille, France. European Language Resources
  Association.

\bibitem[{Laranjo et~al.(2018)Laranjo, Dunn, Tong, Kocaballi, Chen, Bashir,
  Surian, Gallego, Magrabi, Lau, and Coiera}]{laranjo2018conversational}
Liliana Laranjo, Adam~G Dunn, Huong~Ly Tong, Ahmet~Baki Kocaballi, Jessica
  Chen, Rabia Bashir, Didi Surian, Blanca Gallego, Farah Magrabi, Annie Y~S
  Lau, and Enrico Coiera. 2018.
\newblock \href
  {https://academic.oup.com/jamia/article-pdf/25/9/1248/25643433/ocy072.pdf}
  {Conversational agents in healthcare: a systematic review}.
\newblock \emph{Journal of the American Medical Informatics Association},
  25(9):1248--1258.

\bibitem[{Li et~al.(2023)Li, Li, Zhang, Dan, and
  Zhang}]{yunxiang2023chatdoctor}
Yunxiang Li, Zihan Li, Kai Zhang, Ruilong Dan, and You Zhang. 2023.
\newblock \href {http://arxiv.org/abs/2303.14070} {Chatdoctor: A medical chat
  model fine-tuned on llama model using medical domain knowledge}.

\bibitem[{Liu et~al.(2021)Liu, Wan, He, Peng, and Yu}]{liu2021kgbart}
Ye~Liu, Yao Wan, Lifang He, Hao Peng, and Philip~S. Yu. 2021.
\newblock \href {http://arxiv.org/abs/2009.12677} {Kg-bart: Knowledge
  graph-augmented bart for generative commonsense reasoning}.

\bibitem[{Llanos et~al.(2015)Llanos, Bouamor, Bilinski, Ligozat, Zweigenbaum,
  and Rosset}]{llanos2015description}
Leonardo~Campillos Llanos, Dhouha Bouamor, {\'E}ric Bilinski, Anne-Laure
  Ligozat, Pierre Zweigenbaum, and Sophie Rosset. 2015.
\newblock Description of the patientgenesys dialogue system.
\newblock In \emph{Proceedings of the 16th Annual Meeting of the Special
  Interest Group on Discourse and Dialogue}, pages 438--440, Prague, Czech
  Republic. Association for Computational Linguistics.

\bibitem[{Lucke(2023)}]{lucke2023}
Michael Lucke. 2023.
\newblock Using text-based representations of knowledge graphs to improve the
  consistency of generated text.
\newblock Retrieved from the University of Minnesota Digital Conservancy,
  \url{https://hdl.handle.net/11299/253403}.

\bibitem[{Mangrulkar et~al.(2022)Mangrulkar, Gugger, Debut, Belkada, and
  Paul}]{peft}
Sourab Mangrulkar, Sylvain Gugger, Lysandre Debut, Younes Belkada, and Sayak
  Paul. 2022.
\newblock Peft: State-of-the-art parameter-efficient fine-tuning methods.
\newblock \url{https://github.com/huggingface/peft}.

\bibitem[{Nan et~al.(2021)Nan, Nogueira~dos Santos, Zhu, Ng, McKeown,
  Nallapati, Zhang, Wang, Arnold, and Xiang}]{nan-etal-2021-improving}
Feng Nan, Cicero Nogueira~dos Santos, Henghui Zhu, Patrick Ng, Kathleen
  McKeown, Ramesh Nallapati, Dejiao Zhang, Zhiguo Wang, Andrew~O. Arnold, and
  Bing Xiang. 2021.
\newblock \href {https://doi.org/10.18653/v1/2021.acl-long.536} {Improving
  factual consistency of abstractive summarization via question answering}.
\newblock In \emph{Proceedings of the 59th Annual Meeting of the Association
  for Computational Linguistics and the 11th International Joint Conference on
  Natural Language Processing (Volume 1: Long Papers)}, pages 6881--6894,
  Online. Association for Computational Linguistics.

\bibitem[{Nirenburg et~al.(2008)Nirenburg, Beale, McShane, Jarrell, and
  Fantry}]{nirenburg2008language}
Sergei Nirenburg, Stephen Beale, Marjorie McShane, Bruce Jarrell, and George
  Fantry. 2008.
\newblock Language understanding in maryland virtual patient.
\newblock In \emph{Coling 2008: Proceedings of the workshop on Speech
  Processing for Safety Critical Translation and Pervasive Applications}, pages
  36--39, Manchester, UK. Coling 2008 Organizing Committee.

\bibitem[{Ouyang et~al.(2022)Ouyang, Wu, Jiang, Almeida, Wainwright, Mishkin,
  Zhang, Agarwal, Slama, Ray, Schulman, Hilton, Kelton, Miller, Simens, Askell,
  Welinder, Christiano, Leike, and Lowe}]{ouyang2022training}
Long Ouyang, Jeff Wu, Xu~Jiang, Diogo Almeida, Carroll~L. Wainwright, Pamela
  Mishkin, Chong Zhang, Sandhini Agarwal, Katarina Slama, Alex Ray, John
  Schulman, Jacob Hilton, Fraser Kelton, Luke Miller, Maddie Simens, Amanda
  Askell, Peter Welinder, Paul Christiano, Jan Leike, and Ryan Lowe. 2022.
\newblock \href {http://arxiv.org/abs/2203.02155} {Training language models to
  follow instructions with human feedback}.

\bibitem[{Petroni et~al.(2019)Petroni, Rockt{\"a}schel, Riedel, Lewis, Bakhtin,
  Wu, and Miller}]{petroni-etal-2019-language}
Fabio Petroni, Tim Rockt{\"a}schel, Sebastian Riedel, Patrick Lewis, Anton
  Bakhtin, Yuxiang Wu, and Alexander Miller. 2019.
\newblock \href {https://doi.org/10.18653/v1/D19-1250} {Language models as
  knowledge bases?}
\newblock In \emph{Proceedings of the 2019 Conference on Empirical Methods in
  Natural Language Processing and the 9th International Joint Conference on
  Natural Language Processing (EMNLP-IJCNLP)}, pages 2463--2473, Hong Kong,
  China. Association for Computational Linguistics.

\bibitem[{Rashkin et~al.(2022)Rashkin, Nikolaev, Lamm, Aroyo, Collins, Das,
  Petrov, Tomar, Turc, and Reitter}]{rashkin2022measuring}
Hannah Rashkin, Vitaly Nikolaev, Matthew Lamm, Lora Aroyo, Michael Collins,
  Dipanjan Das, Slav Petrov, Gaurav~Singh Tomar, Iulia Turc, and David Reitter.
  2022.
\newblock \href {http://arxiv.org/abs/2112.12870} {Measuring attribution in
  natural language generation models}.

\bibitem[{Sanh et~al.(2019)Sanh, Debut, Chaumond, and
  Wolf}]{sanh2019distilbert}
Victor Sanh, Lysandre Debut, Julien Chaumond, and Thomas Wolf. 2019.
\newblock \href {http://arxiv.org/abs/1910.01108} {Distilbert, a distilled
  version of {BERT:} smaller, faster, cheaper and lighter}.
\newblock \emph{CoRR}, abs/1910.01108.

\bibitem[{Sciuto et~al.(2018)Sciuto, Saini, Forlizzi, and Hong}]{sciuto2018hey}
Alex Sciuto, Arnita Saini, Jodi Forlizzi, and Jason Hong. 2018.
\newblock Hey alexa, what's up?: A mixed-methods studies of in-home
  conversational agent usage.
\newblock In \emph{Proceedings of the 2018 Designing Interactive Systems
  Conference}, pages 857--868.

\bibitem[{Tam et~al.(2022)Tam, Mascarenhas, Zhang, Kwan, Bansal, and
  Raffel}]{tam2022evaluating}
Derek Tam, Anisha Mascarenhas, Shiyue Zhang, Sarah Kwan, Mohit Bansal, and
  Colin Raffel. 2022.
\newblock \href {http://arxiv.org/abs/2211.08412} {Evaluating the factual
  consistency of large language models through summarization}.

\bibitem[{Taori et~al.(2023)Taori, Gulrajani, Zhang, Dubois, Li, Guestrin,
  Liang, and Hashimoto}]{alpaca}
Rohan Taori, Ishaan Gulrajani, Tianyi Zhang, Yann Dubois, Xuechen Li, Carlos
  Guestrin, Percy Liang, and Tatsunori~B. Hashimoto. 2023.
\newblock Stanford alpaca: An instruction-following llama model.
\newblock \url{https://github.com/tatsu-lab/stanford_alpaca}.

\bibitem[{Touvron et~al.(2023)Touvron, Lavril, Izacard, Martinet, Lachaux,
  Lacroix, Rozière, Goyal, Hambro, Azhar, Rodriguez, Joulin, Grave, and
  Lample}]{touvron2023llama}
Hugo Touvron, Thibaut Lavril, Gautier Izacard, Xavier Martinet, Marie-Anne
  Lachaux, Timothée Lacroix, Baptiste Rozière, Naman Goyal, Eric Hambro,
  Faisal Azhar, Aurelien Rodriguez, Armand Joulin, Edouard Grave, and Guillaume
  Lample. 2023.
\newblock \href {http://arxiv.org/abs/2302.13971} {Llama: Open and efficient
  foundation language models}.

\bibitem[{Wang et~al.(2020)Wang, Liu, and Song}]{wang2020language}
Chenguang Wang, Xiao Liu, and Dawn Song. 2020.
\newblock \href {http://arxiv.org/abs/2010.11967} {Language models are open
  knowledge graphs}.

\bibitem[{Wang et~al.(2022)Wang, Kordi, Mishra, Liu, Smith, Khashabi, and
  Hajishirzi}]{wang2022selfinstruct}
Yizhong Wang, Yeganeh Kordi, Swaroop Mishra, Alisa Liu, Noah~A. Smith, Daniel
  Khashabi, and Hannaneh Hajishirzi. 2022.
\newblock \href {http://arxiv.org/abs/2212.10560} {Self-instruct: Aligning
  language model with self generated instructions}.

\bibitem[{Zhang et~al.(2019)Zhang, Zhao, Saleh, and Liu}]{zhang2019pegasus}
Jingqing Zhang, Yao Zhao, Mohammad Saleh, and Peter~J. Liu. 2019.
\newblock \href {http://arxiv.org/abs/1912.08777} {Pegasus: Pre-training with
  extracted gap-sentences for abstractive summarization}.

\end{thebibliography}
\bibliographystyle{acl_natbib}

\appendix
\label{sec:appendix}

\section{Domains of Interest}\label{aa0}

1. dressing \: 2. grooming \: 3. bathing \:4. toileting \:5. incontinence accident management \:6. housekeeping light \:7. housekeeping heavy \:8. laundry \:9. finance \:10. food consumption \:11. meal preparation \:12. meal planing \:13. mobility \:14. transfer \:15. mode of transfer \:16. positioning \:17. mode of positioning \:18. fine motor skills

\section{Example Labeled Survey Form}\label{es}
\begin{table}[H]
\centering
\begin{tabularx}{0.9\linewidth}{|l|L|}
\hline
\textbf{Domain} & Grooming \\
\textbf{Ability} & Physical Assistance \\
\textbf{Age} & 65-84 \\ 
\textbf{Gender} & Female \\ 
\textbf{Assessor-1} & "Can you tell me about how you take care of your grooming needs?" \textit{intent - generic} \\
\textbf{Participant-1} & "I have a hard time" \\ 
\textbf{Assessor-2} & "Can you brush your hair?"\textit{intent - challenges}  \\
\textbf{Participant-2} & "No, I can’t reach my hair to get it brushed in the back." \\
\textbf{Assessor-3} & "Who helps you to brush your hair?"\textit{intent - helper}  \\
\textbf{Participant-2} & "My daughter helps me to brush my hair."\\
\hline
\end{tabularx}
\caption{Sample dialogue from the survey.}
\label{sample survey}
\end{table}

\section{Profile Information}\label{aa1}
\begin{table}[H]
\centering
\begin{tabularx}{\linewidth}{c|c|c|L|L}
\hline
\textbf{\fontsize{9}{9}\selectfont ID} & \textbf{\fontsize{9}{9}\selectfont Age} & \textbf{\fontsize{9}{9}\selectfont Gender} &\textbf{\fontsize{9}{9}\selectfont Avg rating} & \textbf{\fontsize{9}{9}\selectfont \#utterances} \\
\hline
3b1 & 27 & Female & 3.41 & 252\\
3b108 & 64 & Male & 2.73 & 259\\
3b77 & 71 & Female & 3.23 & 196\\
3b84 & 84 & Male & 2.57 & 148\\
3b86 & 52 & Male & 3.53 & 206\\
4d18 & 86 & Female & 3.58 & 233\\
4d23 & 60 & Male & 3.78 & 114\\
4d26 & 96 & Female & 3.54 & 81\\
4d29 & 42 & Female & 1.74 & 50\\
4d4 & 63 & Female & 3.07 & 213\\
\hline

\end{tabularx}
\caption{Synthetic profile summary}
\label{profiles}
\end{table}



\section{Evaluation Questions}

\subsection{Test questions}\label{aa4}

\noindent
\textbf{Bathing}
\begin{enumerate}
    \item Tell me about how bathing goes for you.
    \item Can you elaborate more on that?
    \item Can you get in and out of the shower easily?
    \item Do you need any help with drying off?
    \item Can you wash your back okay?
\end{enumerate}
\noindent
\textbf{Dressing}
\begin{enumerate}
    \item Tell me about how you get dressed in the morning.
    \item Is there anything else I should know about that?
    \item What about buttons and zippers specifically? Do you struggle at all with them?
    \item Can you manage your shoes on your own?
    \item Do you prefer any particular type of clothing?
\end{enumerate}

\subsection{Example of long casual talks for factuality evaluation}

\textbf{Bathing assessment}

[Assessor]: Tell me about how bathing goes for you?

[Chatbot]: I have to get my wife or a caregiver to help me bathe. I can't do it myself.

[Assessor]: What sort of help does your wife give you?

[Chatbot]: She helps me undress, wash, dry off, dress, and put on my clothes. She also helps me shave.

[Assessor]: What parts of washing do you struggle with?

[Chatbot]: I can't reach my back.

[Assessor]: But you can manage to turn on the water yourself?

[Chatbot]: Yes, but I can't reach the soap.

[Assessor]: Have you thought about getting a caddy or something to bring the soap closer to you?

[Chatbot]: No, I haven't.

[Assessor]: Okay. Good to know. Is it easy to dry off after you shower?

[Chatbot]: It is hard to dry off because I am not able to stand up by myself.

[Assessor]: How long does it take you to shower and how often do you get to do it?

[Chatbot]: I don't like to take too much time in the shower because I get cold easily. It takes me about 15 minutes to shower.

[Assessor]: How frequently are you able to shower?

[Chatbot]: I am able to shower every other day.

[Assessor]: Do you prefer when your caregiver helps? Or your wife?

[Chatbot]: My wife helps me.

[Assessor]: Do you have any grab bars or anti-slip mats or other things that help you stay steady in the shower?

[Chatbot]: I have a bench in the shower.

\section{Domain F1 score}
\begin{figure}[htbp]
  \centering
    \includegraphics[width=\linewidth]{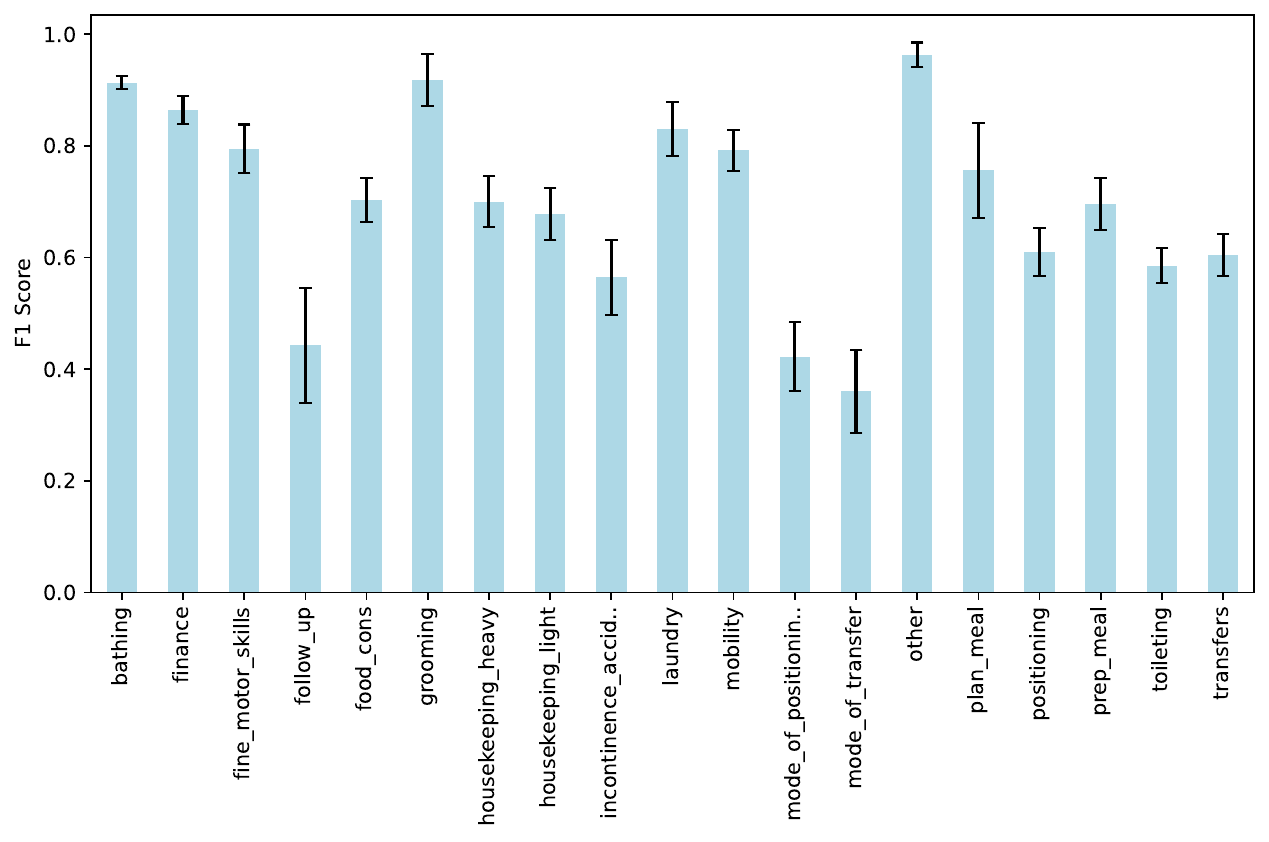}
    \caption{F1 score for each domain, aggregated across experiments. (Dressing domain is not included because we didn't derive the test data. And the results for bathing domain is from augmented set only as the original set does not have test data either.)}
    \label{domainf1}
\end{figure}

\end{document}